\documentclass[11pt]{article}
\pdfoutput=1

\usepackage[]{EACL2023}

\usepackage{times}
\usepackage{latexsym}
\usepackage{url}
\usepackage{hyperref}
\usepackage{subcaption}
\usepackage{arydshln}
\usepackage[T1]{fontenc}

\usepackage[utf8]{inputenc}

\usepackage{microtype}
\usepackage{subcaption}
\usepackage{inconsolata}
\usepackage{graphicx}
\usepackage{array, booktabs, tablefootnote, float}
\definecolor{esc}{RGB}{0, 153, 51}

\title{Equipping Language Models with Tool Use Capability\\ for Tabular Data Analysis in Finance}

\author{
  Adrian Theuma\ \ \ \ \ 
  Ehsan Shareghi\\
  Department of Data Science \& AI, Monash University \\
  {adriantheuma@gmail.com}\ \ \ \ \ \ {ehsan.shareghi@monash.edu}
}

\begin{document}
\maketitle

\begin{abstract}

Large language models (LLMs) have exhibited an array of reasoning capabilities but face challenges like error propagation and hallucination, particularly in specialised areas like finance, where data is heterogeneous, and precision is paramount. We explore the potential of language model augmentation with external tools to mitigate these limitations and offload certain reasoning steps to external tools that are more suited for the task, instead of solely depending on the LLM's inherent abilities. More concretely, using financial domain question-answering datasets, we apply supervised fine-tuning on a \textsc{LLaMA-2 13B Chat} model to act both as a \emph{task router} and \emph{task solver}. The \emph{task router} dynamically directs a question to either be answered internally by the LLM or externally via the right tool from the tool set. Our tool-equipped SFT model, \textsc{Raven}, demonstrates an improvement of $35.2\%$ and $5.06\%$ over the base model and SFT-only baselines, respectively, and is highly competitive with strong GPT-3.5 results. To the best of our knowledge, our work is the first that investigates tool augmentation of language models for the finance domain.\footnote{Code, model, and data: \url{https://raven-lm.github.io}}
%
%

\end{abstract}

\section{Introduction}\label{sec:intro}
Augmenting Large Language Models (LLMs) with tools has emerged as a promising approach to further complement LLMs' capabilities with  specialised mechanisms, leading to improved accuracy and reliability~\cite{DBLP:journals/corr/abs-2302-04761, DBLP:conf/iclr/YaoZYDSN023}. 
%
%
%
%
%
This approach offloads tasks, fully or partially, to a deterministic offline tool such as a python interpreter \citep{DBLP:conf/icml/GaoMZ00YCN23}, calculator \citep{DBLP:journals/corr/abs-2110-14168}, knowledge base~\citep{DBLP:conf/icml/BorgeaudMHCRM0L22}, or online APIs of models and services~\cite{DBLP:conf/iclr/YaoZYDSN023, qin2023toolllm, shen2023hugginggpt}. 

This paradigm holds particular appeal in fields demanding precision, such as finance~\citep{DBLP:journals/corr/abs-2306-06031} and healthcare \citep{DBLP:journals/bib/LuoSXQZPL22, DBLP:journals/corr/abs-2212-13138}. 
Specifically, the specialised terminology within the finance domain and the diverse range of data sources, encompassing both structured and unstructured data, along with the complex numerical reasoning requirements across such heterogeneous sources, render it an ideal candidate for potential improvements through tool augmentation. Nevertheless, there has been limited research dedicated to this specialised domain.

A satisfying review of existing works on tool augmentation of LLMs is beyond the scope of this work; however, this space can be divided into two primary directions: (1) approaches that require an LLM at the center and uses few-shot in-context learning to either provide tool and API documentations, or demonstrations that involve tool use~\cite{DBLP:journals/corr/abs-2308-00675, qin2023toolllm, shen2023hugginggpt,DBLP:journals/corr/abs-2308-00675},  and (2) approaches that build fine-tuned smaller LMs under a static tool use protocol~\cite{DBLP:journals/corr/abs-2302-04761}, or through expensive annotations collected from commercial LLMs~\cite{chen2023fireact, DBLP:conf/iclr/YaoZYDSN023}. 

\begin{figure*}[ht]
\centering
\includegraphics[trim={0.5cm 1.9cm 0.5cm 2.8cm},clip, width=1\textwidth]{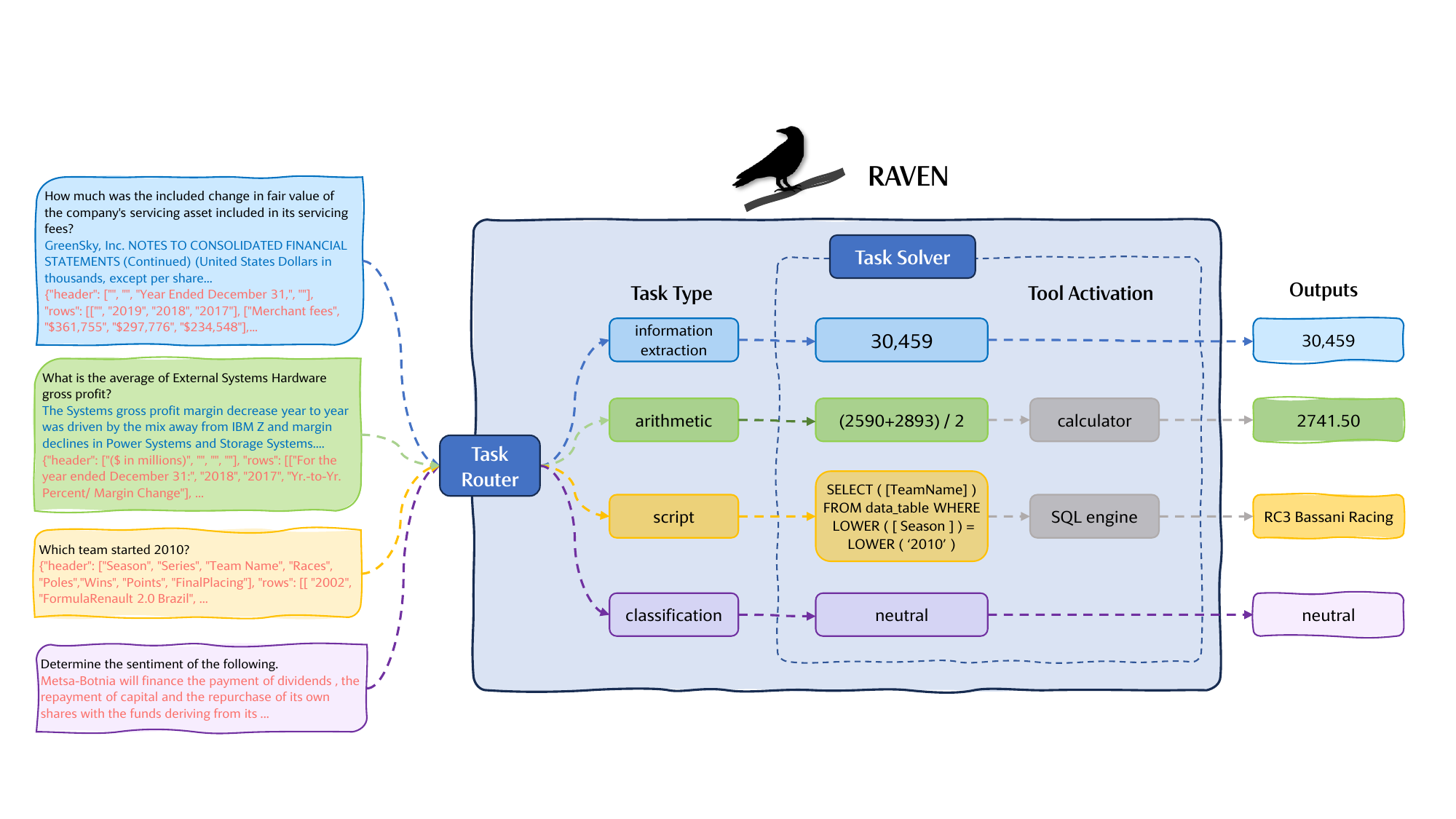}
\caption{\textbf{\textsc{Raven} Inference Flow}. Using the language model the \textit{Task Router} infers the optimal task format to use conditioned on the given prompt. The \textit{Task Solver} re-formats the instruction according to the selected template by the task router and sends it to the language model again. The pipeline will branch between serving the response directly or calling a tool API to perform an intermediate evaluation before serving the final output.} \label{fig:inference}
\vspace{-2mm}
\end{figure*}



In this work, our primary focus lies in demonstrating the potential of tool augmentation within the finance domain. Acknowledging the utmost significance of privacy concerns within the financial sector, we have chosen to adopt a fully offline approach, equipping a language model with diverse tool utilisation mechanisms. More concretely, we employ Parameter Efficient Fine-Tuning (PEFT)~\citep{DBLP:conf/iclr/HuSWALWWC22,DBLP:conf/icml/HoulsbyGJMLGAG19} to equip a \textsc{Llama 2 13B chat} \citep{DBLP:journals/corr/abs-2307-09288} with tool use capabilities. 
Our approach differs from previous research in two significant ways. First, we do not rely on costly annotations of training examples produced by commercial language models. Second, we enhance existing question-answering training datasets by incorporating instructions and merge data representing various tasks. This approach instructs the model to adapt dynamically and determine the most appropriate mechanism (either internal or tool-based) to address each specific query. 

Our model, \textsc{Raven}, achieves significant improvements in reasoning over structured data. For example, compared to the base model we demonstrate a lift in exact match accuracy of $ 63.8\%$ ($ 21.68\%  \rightarrow 85.52\% $) on the \textsc{Wiki-SQL}~\citep{DBLP:conf/emnlp/XuWWFS18}. Despite being much smaller in size, \textsc{Raven} also outperforms GPT-3.5 on all benchmarks with an absolute average accuracy lift of $ 9.2\% $.    

\section{\textsc{Raven}}\label{sec:raven}
We use the \textsc{Llama 2 13B chat} \citep{DBLP:journals/corr/abs-2307-09288} model as the backbone and fine-tune it using Low Rank Adaptation (\textsc{LoRA}) \citep{DBLP:conf/iclr/HuSWALWWC22}. In this section we provide training details of \textsc{Raven}. The overall architecture of \textsc{Raven} is shown in Figure~\ref{fig:inference}.

\subsection{Fine-tuning Data}\label{subsec:raven_dataprep}
We use a mixture of four financial and generic structured and unstructured question-answering datasets. 
We provide a brief summary in below. 

\vspace{2mm}
\noindent\textbf{TAT-QA}. Consists of questions generated by financial experts associated with 
hybrid contexts drawn from real-world financial reports~\citep{DBLP:conf/acl/ZhuLHWZLFC20}. The questions typically require a range of data extraction and numerical reasoning skills, including multiplication, comparison, sorting, and their various combinations. Apart from the answer, TAT-QA also provides the derivation, where applicable, which proves beneficial for offloading the calculation to an external tool, as will be explained in \S\ref{subsec:raven_tools}. 

\vspace{2mm}
\noindent\textbf{Financial PhraseBank}. Consists of phrases derived from English news on listed companies in OMX Helsinki \citep{DBLP:journals/jasis/MaloSKWT14}. The dataset contains phrase-level annotation by financial markets experts, that categorise each sample sentence as either positive, negative, or neutral, from an investor's standpoint. This dataset is relevant because sentiment analysis models trained on general datasets do not perform well in specialised domains due to the unique vocabulary found in financial texts, which often do not rely on easily identifiable positive or negative words~\citep{DBLP:journals/corr/abs-1908-10063}. 

\vspace{2mm}
\noindent\textbf{Wiki-SQL}. Consists of manually annotated crowd sourced examples of natural language questions and SQL queries over tables found on Wikipedia \citep{DBLP:journals/corr/abs-1709-00103}. Whilst this is not specifically a financial domain dataset its relevancy is in the availability of the script that produces the answer. Similar to the derivation in the TAT-QA dataset this script is crucial to steer our model to use a tool instead of producing the answer directly. 

\vspace{2mm}
\noindent\textbf{OTT-QA}. Similar to \textsc{TAT-QA}, this dataset consists of questions over tabular data and unstructured text across diverse domains \citep{DBLP:conf/iclr/ChenCSWC21}. The majority of questions necessitate multi-hop inference involving both forms of data. The dataset's relevance lies in its omission of derivation or intermediate steps, which poses a challenge for the model to infer the correct answer.

\vspace{2mm}   
\noindent\textbf{Data splits.} 
Among the four datasets, FPB\footnote{\url{https://github.com/vrunm/Text-Classification-Financial-Phrase-Bank}} and OTT-QA\footnote{\url{https://github.com/wenhuchen/OTT-QA}} lack a published test split. TAT-QA\footnote{\url{https://nextplusplus.github.io/TAT-QA/}} has a test split without gold labels. WikiSQL\footnote{\url{https://github.com/salesforce/WikiSQL}} provides a public test set. We used the WikiSQL test split, and for the other 3 datasets generated random 80-10-10 splits~(available \href{https://huggingface.co/datasets/adriantheuma/raven-data}{here}). Table \ref{tab:datasets} summarises the statistics of the datasets. 

  \begin{table*}[t]
    \centering
    \resizebox{\textwidth}{!}{%
    \begin{tabular}{ l c c c c c c c c c c }
    \toprule
        & \multicolumn{3}{c}{\textsc{Statistics}}&\multicolumn{7}{c}{\textsc{Models}} \\ \cmidrule(lr){2-4}\cmidrule(lr){5-11}
                 Dataset& Train  & Dev  & Test & \textsc{GPT-3.5 (CoT)} & \textsc{GPT-3.5 (5-shot)} & \textsc{+Tools} & \textsc{Llama2}& \textsc{+SFT} & \textsc{Raven}  & \textsc{+Backoff}  \\ \cmidrule(lr){1-1}\cmidrule(lr){2-4}\cmidrule(lr){5-7}\cmidrule(lr){8-9}\cmidrule(lr){10-11}
        TAT-QA   & 10,477 & 1,162  & 1,278 & 19.23\% & 34.06\% & 46.82\% & 10.91\% & 37.87\% & 51.35\% & \textbf{52.27\% }    \\
        OTT-QA   & 10,273 & 1,115 & 1,247 & 5.55\% & 14.55\%  & 14.60\% & 6.18\% & \textbf{20.10\%} & 16.03\% & 16.03\%   \\
        Wiki-SQL  & 12,782 & 1,391 & 1,536 & 32.07\% & 53.00\% & 75.88\% & 21.68\% & 74.38\% & 84.25\% & \textbf{85.52\%}    \\
        FPB      & 3,413 & 382 & 421 & 44.18\% & 70.07\% & 71.73\% & 66.03\% & 90.97\% & \textbf{91.92}\% & 91.92\%       \\\bottomrule
    \end{tabular}
    }
    \caption{The data statistics and experimental results (Exact Match) of various benchmarks and models. The best results are in \textbf{bold}. GPT-3.5 results are based on 5-shots. \textsc{Sota} is based on previously published results.}
    \label{tab:datasets}
\end{table*}
\subsection{Tools}\label{subsec:raven_tools}
\textsc{Raven} is equipped with two external offline tools: a calculator and a SQL engine. The \emph{Calculator} is instantiated in a python interpreter and is used to evaluate well-formed arithmetic expressions. The API expects one input representing the arithmetic expression and returns the evaluated result. The \emph{Lightweight SQL engine} is an API capable of executing SQL scripts on relational data. The API expects two inputs, (1) a string representation of the structured data and (2) a SQL script. The API's lightweight database engine converts structured data from its textual form to the engine's relational representation and converts data types where applicable. The SQL script is executed on this representation and the API returns the result.

\subsection{Instruction Tuning}\label{subsec:raven_tuning}
Inspired by \citet{DBLP:conf/acl/WangKMLSKH23} and \citet{alpaca} we engineer various templates for SFT instruction tuning.
%
In general, we require to extract up to four key attributes from the original datasets. These are (1) \textit{instruction} that describes the task to perform, for example, "\textit{Determine the sentiment of the following phrase}", or the question "\textit{What is the percentage change in revenue after the adoption of ASC 606?}" (2) \textit{input} that provides more context such as the phrase to classify or a passage, (3) \textit{data} that accompanies the context in tabular format, (4) \textit{derivation} that produces the answer or expected \textit{response}. The instruction and one of derivation or response are mandatory, whilst the other attributes are included if applicable.

To ensure training diversity, our model is trained on a combination of all available training data. Based on the data, we craft different templates depending on which tool the model should choose or if the model should directly answer the question on its own (i.e., to train the \emph{Task Solver} in Figure~\ref{fig:inference}). We also automatically generate another dataset, that supplements the above question-answer dataset for training our model to select the appropriate template based on the context (i.e., to train the \emph{Task Router} in  Figure~\ref{fig:inference}). Refer to appendix \ref{app:templates} for template examples.

\subsection{Inference}
During inference, we follow a two-step process with \textsc{Raven}. First, we employ a specialised \textit{template choice} prompt to determine the most suitable prompt template (from "arithmetic," "classification," "script," or "information extraction") based on the input. Next, we wrap the instruction, including the input and relevant data, in the inferred prompt template and send it to \textsc{Raven} for generating the subsequent output. Depending on the selected template, the \textit{Task Solver} either activates a tool to fulfil the request or directly produces the response.
%
%

We discuss the inference behaviour when each of these templates are used. For \textbf{Script} the model is expected to produce a well-structured SQL script. In this scenario, the structured data table provided in the prompt is temporarily loaded in memory using a lightweight database engine, and the script execution on the table produces the output. For \textbf{Arithmetic} the model is expected to predict a well formed arithmetic expression. This expression is evaluated by a calculator and the resulting value passed as output. The \textbf{Information Extraction} template instructs the model that there is information included in structured form that needs to be considered before producing the answer. In this case no tool is used and the model is expected to infer the correct output based solely on the information in the prompt. The \textbf{Classification} template is used when the prediction of the model should be taken as-is. 

\section{Experiments}\label{sec:experiments}
We compare with the base \textsc{Llama 2 13B Chat} with and without SFT\footnote{To steer the base model into producing a short answer we add \textit{"Output the answer only with no explanation."} to the prompt.}. We also report GPT-3.5\footnote{gpt-3.5-turbo} (5-shot), GPT-3.5 (Chain-of-Thought~\cite{DBLP:conf/nips/Wei0SBIXCLZ22}) and GPT-3.5 (5-shot + Tools).
The SFT model trained with tool use is denoted as \textsc{Raven}. When tool use fails due to ill-formed arguments we have a fallback mechanism to produce the answer by the SFT model, denoted as \textsc{Backoff}. For training details and hardware, see Appendix~\ref{app:training}. We evaluate the models using \textit{exact match}. The task router has determined the correct type 100\% of the time, except for \textsc{TAT-QA} where the accuracy was 90.62\%.

\subsection{Main Results}\label{sec:results}
The results are summarised in Table \ref{tab:datasets}. 
%
%
Compared to the base model, \textsc{Raven} significantly improves the results on the \textbf{PhraseBank} dataset by an absolute 25.9\%. 
%
%
On the \textbf{Wiki-SQL} dataset the base model is \emph{unable} to infer the correct answer almost 80\% of the time. This figure is inverted
for \textsc{Raven} which obtains a \emph{4-fold} improvement over the base model inferring the correct answer more than 85\% of the time. Our model improves on the best \textsc{GPT-3.5} performance by close to 10\% (absolute). All the questions in this dataset can be addressed using the lightweight database engine and involve a combination of data selection, ranking and arithmetic operations on structured data. This result underscores the distinct advantage of delegating this task to a tool rather than relying on the language model to infer the results in a zero-shot manner. Despite the results not being as strong as \textsc{Raven} we observe a similar pattern on the \textsc{GPT-3.5} evaluation in which better results are incrementally obtained when including examples in the context and using tools compared to \textsc{CoT}. 

We see a similar pattern on the \textbf{TAT-QA} benchmark with the tool augmented model achieving a \emph{5-fold} improvement on the base model. Approximately 46\% of the observations of the \textsc{TAT-QA} dataset are annotated with an intermediate arithmetic derivation that \textsc{Raven} evaluates using a calculator at inference time. We perform a comparative analysis to explore whether our model performs better on this portion of the data in the analysis section (\S\ref{subsec:results_analysis}).  

In \textbf{OTT-QA}, the majority of questions require multi-hop inference involving both tabular data and unstructured text, with the information needed to answer the questions dispersed across these two input types. 
This dataset does not have annotated intermediate steps to get to the answer and therefore all models are expected to infer the answer without relying on tools. Despite \textsc{SFT} achieving an increase in accuracy compared to the base model, the relatively low score underscores the importance of intermediate reasoning steps and tools~\cite{chen2023fireact}. 

We observed the \textsc{Backoff} mechanism to bring slight improvement on \textsc{TAT-QA} ($51.35\% \rightarrow 52.27\%$) and \textsc{Wiki-SQL} ($84.25\% \rightarrow 85.52\%$).

\begin{figure}[!ht]
    \centering
    \includegraphics[
        width=10cm,
        height=5cm, 
        keepaspectratio
        ]{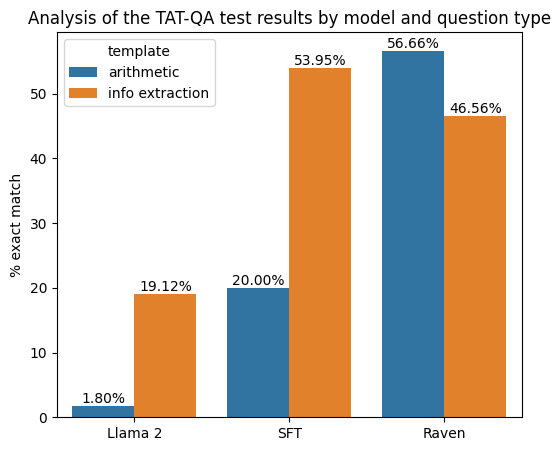}
    \caption{Comparison of model performance on the \textsc{TAT-QA} dataset specifically highlighting the effect of a tools-augmented model on questions that require multi-hop reasoning.}
    \label{fig:tat_qa_results}
\end{figure}

\begin{figure}[!ht]
    \centering
    \includegraphics[
        width=10cm,
        height=5cm, 
        keepaspectratio
        ]{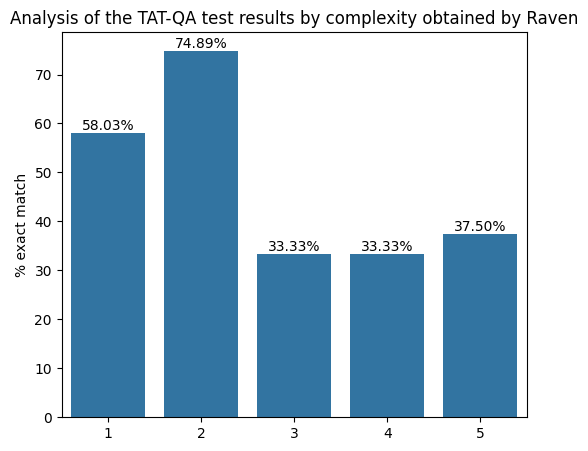}
    \caption{Comparison of model performance on the \textsc{TAT-QA} dataset highlighting the effect of complexity on model performance.}
    \label{fig:tat_qa_results_complexity}
\end{figure}

\subsection{Analysis}\label{subsec:results_analysis}
\vspace{1mm}
\noindent\textbf{Is it better to have a separate model for each task?} 
We developed a model specifically using the \textsc{TAT-QA} dataset, achieving an evaluation score of 54.70\%. This dedicated model outperforms \textsc{Raven} by 2.4\%. We contend that this modest performance gain does not warrant the added complexity of maintaining separate models and switching between them during inference.

\vspace{2mm}
\noindent\textbf{Why tool augmentation is necessary?} Approximately half of the questions within the \textsc{TAT-QA} dataset are annotated with an arithmetic equation. The presence of the equation implies that the language model needs to perform multiple actions to output the correct answer. This process involves the correct extraction of, at a minimum, two numerical values from the context, followed by the execution of an arithmetic operation, such as addition or division. This particular scenario is ideal to understand the effect of SFT and tool augmentation by comparing the performance of different models on the two categories of data from the same dataset. As shown in Figure \ref{fig:tat_qa_results} the base model without any fine-tuning is ill-equipped to perform multi-hop reasoning achieving close to 2\% accuracy equating to ten correct answers of approximately 620. Although we observe an improvement in the SFT model, the impact of using tools is evident in the substantial jump to 56.7\% accuracy achieved by \textsc{Raven}. These findings further confirm that SFT models are able to accurately extract multiple data points from the context but require external tools to correctly compose the final answer from the gathered data. This is also evidenced by the consistent performance of the \emph{Information Extraction} type questions between \textsc{SFT} and \textsc{Raven}, which only requires data extraction to answer the question. 

The utility of augmenting language models with external tools is substantiated further through a comparative analysis of experimental outcomes on two similar datasets. Addressing questions on \textsc{Wiki-SQL} and \textsc{OTT-QA} requires multi-hop reasoning across diverse forms of data, spanning both structured and unstructured formats. The primary difference lies in the annotation method: the \textsc{Wiki-SQL} dataset is annotated with a data extraction script which, when executed on the structured data, yields the answer. In contrast, the \textsc{OTT-QA} dataset lacks this intermediate derivation step. By delegating the script execution to an external tool, \textsc{Raven} achieves an exact match accuracy of 85.52\% on \textsc{Wiki-SQL} and 16.03\% on \textsc{OTT-QA}, underscoring the effectiveness of fit-for-purpose external tools in this scenario. 

\vspace{2mm}
\noindent\textbf{What is the impact of question complexity?} On the \textsc{TAT-QA} dataset we can use the number of arithmetic operators in the \emph{gold} arithmetic equation as a proxy for question complexity. One arithmetic operator implies the extraction of two numerical values from the context, two operators, three numerical values, and so on. As shown in Figure \ref{fig:tat_qa_results}, \textsc{Raven}'s performance degrades with the number of numerical values to extract from the context. 

%
\section{Conclusion}
In this paper we have demonstrated the feasibility of equipping a \textsc{Llama 2 13B chat} model with tool use capabilities via fine-tuning a mere 0.2\% of its parameters on a relatively small and diverse dataset. The augmentation with tools remarkably elevated the performance of the base model by an average of 35.2\% across 4 datasets, surpassing even a significantly larger GPT-3.5 model by 9.2\%. Additionally, through a comparative analysis
of question answering datasets we demonstrate the effectiveness of augmenting language models with external tools, showing significant improvements in accuracy when addressing multi-hop questions with tools.


\section*{Limitations} 
\noindent\textbf{Infrastructure Bottleneck}. Our experiments were constrained with fitting our model on available commodity hardware. We hypothesise that it would be possible to obtain better performance using the larger \textsc{Llama 2} 70 billion-parameter model and a longer context length. Experiments by \citet{DBLP:journals/corr/abs-2307-09288} demonstrated that the 70-billion-parameter model consistently achieves the highest performance across various prominent natural language understanding benchmarks. Additionally, a longer context length enables experimentation with diverse prompts as well as alternative representations of structured data.

\vspace{2mm}
\noindent\textbf{Language model evaluation}.
Free-form natural language generation (NLG) poses significant evaluation challenges that remain under-studied to this date \citep{DBLP:journals/corr/abs-2303-16634}. \citet{DBLP:journals/corr/abs-2306-05685} argue that while users prefer the responses of an instruction-tuned model over the base model, traditional LLM benchmarks \citep{DBLP:journals/corr/abs-2211-09110,DBLP:conf/iclr/HendrycksBBZMSS21} cannot tell the difference. This challenge is heightened in specialised domains such as finance. Common similarity scores such as \textsc{BLEU} \citep{DBLP:conf/acl/PapineniRWZ02} which measures \textit{n-gram} overlap between candidate and reference sentences are unsuitable due to misleading accuracy or penalised semantic correctness \citep{DBLP:conf/wmt/FreitagRMLSAKFLM22}. Although \textsc{BERTScore} \citep{DBLP:conf/iclr/ZhangKWWA20} addresses some of these pitfalls by measuring the similarity of candidate and reference sentences using pre-trained contextualised embeddings it can still produce high scores for inaccurate results. For example the candidate and reference sentences \textit{"The amount of goodwill reallocated to the IOTG operating segment in 2018 was \$480 \textbf{million}"}, and \textit{"The amount of goodwill reallocated to the IOTG operating segment in 2018 was \$480"} have a \textsc{BERTScore} (f1) of 99.17\%! These measures are not suitable for comparing numerical content. 

Conversely, using exact match criteria might unjustly penalise NLG models, given that identical numerical values can be expressed in varying forms - such as "\$4 million" and "\$4,000,000," or "0.24" and "24\%,". In some cases, numerical values can be integrated within a passage of text, rendering the evaluation of such content very challenging. In our evaluation we have normalised different formatting (such as converting values to percentages where appropriate), however a universal normalising algorithm in this space is outside the scope of our research. 

\vspace{2mm}
\noindent\textbf{GPT-3.5 evaluation}. Evaluating our benchmark with \textsc{GPT-3.5} poses significant challenges, especially when using \textsc{Zero-Shot (CoT)} \citep{DBLP:conf/nips/KojimaGRMI22}. \textsc{GPT-3.5} does not consistently adhere to instructions for providing a concise response, such as a single word or number, which makes \emph{exact match} comparisons challenging. Additionally, we have noticed that \textsc{GPT-3.5} does not generate a response when uncertain. This is particularly evident when evaluating the FPB, which does not exhibit common sentiment negative or positive words.

\vspace{2mm}
\section*{Ethics Statement}
Our work is built on top of existing pre-trained language models. Our goal was not to attend to alleviate the well-documented is- sues (e.g., privacy, undesired biases, etc) that such models embody. For this reason, we share the similar potential risks and concerns posed by these models. Additionally, our SFT was conducted on publicly available research benchmarks, and as such the additional SFT step used in \textsc{Raven} is unlikely to introduce any new area of risk.

\bibliography{custom}
\bibliographystyle{acl_natbib}

\appendix

\onecolumn
\section{Background on LMs in Finance}
\citet{DBLP:journals/corr/abs-1908-10063} tackles financial sentiment analysis by further pre-training BERT \citep{DBLP:conf/naacl/DevlinCLT19} on a financial corpus and uses the resulting sentence embeddings to obtain higher text semantic representation before training a downstream sentiment classifier. The author concludes that fine-tuning the generic language model captures the nuances of financial terminology demonstrated by the absolute SOTA improvement of 15\%. \citet{DBLP:journals/corr/abs-2301-04408} use the \textsc{text-davinci-003}\footnote{https://platform.openai.com/docs/models/gpt-3-5} API to assess whether LLMs have the potential to augment industry knowledge workers. In agreement with earlier findings \citep{DBLP:journals/corr/abs-2112-00114}, \citet{DBLP:journals/corr/abs-2301-04408}'s model under-performs human performance by a large margin on quantitative reasoning tasks of the American Institute of Certified Public Accountants (AICPA) assessment while approaching human levels on multiple choice questions, achieving an accuracy rate of  14.4\% and 57.6\% respectively. \citet{DBLP:journals/corr/abs-2303-17564} train a 50 billion parameter LLM using Bloomberg's closed source datasets and general-purpose data to obtain BloombergGPT, the first large scale specialised language model in the finance domain. The resulting model performs well on financial benchmarks while retaining general-purpose performance comparable to other foundational models.

\section{Training Details}\label{app:training}
\noindent\textbf{Training details}. We use the pre-trained weights of \textsc{Llama 2 13B chat} \citep{DBLP:journals/corr/abs-2307-09288} for the base model and \textsc{LlamaTokenizer} for prompt tokenisation. We limit the maximum context length to 1,204 tokens and discard any training observations that exceed this limit after tokenisation. Due to hardware constraints we use a per device train batch of one and accumulate the gradient for 128 steps achieving the equivalent \verb|batch_size| of 128 and use quantisation to load the model in \verb|8-bit| \citep{DBLP:journals/corr/abs-2208-07339}. We adapt the same \verb|optimiser|, \verb|learning_rate| and \verb|warmup_steps| as \citet{alpaca}, and set these to \verb|adamw|, $3\times10^{-4}$ and 100, respectively. We use Low Rank Adaptation to reduce the number of trainable parameters and similar to \citet{alpaca} set the \verb|rank| and \verb|alpha| hyper-parameters to 16, \verb|dropout| to 0.05 and target the \verb|q_proj|, \verb|k_proj|, \verb|v_proj|, and \verb|o_proj| modules of the base model. This reduces the trainable parameters to 26,214,400 or 0.2\% of the base model. The final models are trained for 5 \verb|epochs| totalling 1,200 \verb|steps|.

\vspace{2mm}
\noindent\textbf{Training hardware}. We train the models on commodity hardware equipped with a 13th Gen Intel(R) Core(TM) i7-13700KF CPU at 3.40 GHz, 64 GB installed RAM and NVIDIA GeForce RTX 4090 GPU with 24 GB onboard RAM. The final model consumed 100 GPU hours during training and 10 GPU hours for evaluation. 

\vspace{2mm}
\noindent\textbf{Carbon footprint}. Given we train two models and an average consumption of 400 Wh we estimate the total power consumption to be 88 kWh with a carbon dioxide equivalent (CO\textsubscript{2e}) emissions of 0.081 tonnes\footnote{https://carbonpositiveaustralia.org.au/carbon-footprint-calculator}. To obtain a realistic measure of emissions we also need to consider multiple training experiments with different settings leading to the final models including with different hyper-parameters, prompt templates and other mix of datasets. We estimate the realistic total consumption and emissions is 10-fold that of the final models.

\vspace{2mm}
\noindent\noindent\textbf{GPT-3.5 Experiments}
We compare our results with GPT-3.5 using few-shot in-context learning. We use the following \textit{system} to steer the model into producing a short answer. "\textit{You are a data expert that can reason over structured and unstructured data. Use the following examples to help you reason over the final question. Follow the same format of the examples to answer the final question. Output a short response with the answer only and do not include any explanations or introductory sentences.}"

\section{Templates} \label{app:templates}
Below are a few examples of prompts generated from the datasets used to train \textsc{Raven}.
\subsection{TAT-QA}
\textbf{Example 1 - The response is an equation}\newline
Below is an instruction that describes a task, coupled with input and data providing additional context. Formulate an arithmetic equation to generate the answer.
\\\\
\#\#\# Instruction: \newline What was the change in the basic net earnings per share between 2017 and 2019?
\\\\
\#\#\# Input: \newline (5) Earnings Per Share Basic earnings per share is computed by dividing Net earnings attributable to Black Knight by the weighted-average number of shares of common stock outstanding during the period. For the periods presented, potentially dilutive securities include unvested restricted stock awards and the shares of BKFS Class B common stock prior to the Distribution. For the year ended December 31, 2017, the numerator in the diluted net earnings per share calculation is adjusted to reflect our income tax expense at an expected effective tax rate assuming the conversion of the shares of BKFS Class B common stock into shares of BKFS Class A common stock on a one-for-one basis prior to the Distribution. The effective tax rate for the year ended December 31, 2017 was (16.7)\%, including the effect of the benefit related to the revaluation of our net deferred income tax liability and certain other discrete items recorded during 2017. For the year ended December 31, 2017, the denominator includes approximately 63.1 million shares of BKFS Class B common stock outstanding prior to the Distribution. The denominator also includes the dilutive effect of approximately 0.9 million, 0.6 million and 0.6 million shares of unvested restricted shares of common stock for the years ended December 31, 2019, 2018 and 2017, respectively. The shares of BKFS Class B common stock did not share in the earnings or losses of Black Knight and were, therefore, not participating securities. Accordingly, basic and diluted net earnings per share of BKFS Class B common stock have not been presented. The computation of basic and diluted earnings per share is as follows (in millions, except per share amounts):
\\\\
\#\#\# Data: \newline \{"header": ["", "", "Year ended December 31,", ""], "rows": [["", "2019", "2018", "2017"], ["Basic:", "", "", ""], ["Net earnings attributable to Black Knight", "\$108.8", "\$168.5", "\$182.3"], ["Shares used for basic net earnings per share:", "", "", ""], ["Weighted average shares of common stock outstanding", "147.7", "147.6", "88.7"], ["Basic net earnings per share", "\$0.74", "\$1.14", "\$2.06"], ["Diluted:", "", "", ""], ["Earnings before income taxes and equity in losses of unconsolidated affiliates", "", "", "\$192.4"], ["Income tax benefit excluding the effect of noncontrolling interests", "", "", "(32.2)"], ["Net earnings", "", "", "\$224.6"], ["Net earnings attributable to Black Knight", "\$108.8", "\$168.5", ""], ["Shares used for diluted net earnings per share:", "", "", ""], ["Weighted average shares of common stock outstanding", "147.7", "147.6", "88.7"], ["Dilutive effect of unvested restricted shares of common", "", "", ""], ["stock", "0.9", "0.6", "0.6"], ["Weighted average shares of BKFS Class B common stock outstanding", "", "", "63.1"], ["Weighted average shares of common stock, diluted", "148.6", "148.2", "152.4"], ["Diluted net earnings per share", "\$0.73", "\$1.14", "\$1.47"]]\}
\\\\
\#\#\# Equation: \newline 0.74-2.06 
\\\\
\textbf{Example 2 - The response is determined from the text or table}\newline
Here is a instruction detailing a task, accompanied by input and data providing additional context. Provide a suitable reply that effectively fulfills the inquiry.
\\\\
\#\#\# Instruction: \newline What was the Additions based on tax positions related to current year in 2019 and 2018 respectively?
\\\\
\#\#\# Input: \newline A reconciliation of the beginning and ending amount of unrecognized tax benefits is as follows: Interest and penalty charges, if any, related to uncertain tax positions are classified as income tax expense in the accompanying consolidated statements of operations. As of March 31, 2019 and 2018, the Company had immaterial accrued interest or penalties related to uncertain tax positions. The Company is subject to taxation in the United Kingdom and several foreign jurisdictions. As of March 31, 2019, the Company is no longer subject to examination by taxing authorities in the United Kingdom for years prior to March 31, 2017. The significant foreign jurisdictions in which the Company operates are no longer subject to examination by taxing authorities for years prior to March 31, 2016. In addition, net operating loss carryforwards in certain jurisdictions may be subject to adjustments by taxing authorities in future years when they are utilized. The Company had approximately \$24.9 million of unremitted foreign earnings as of March 31, 2019. Income taxes have been provided on approximately \$10.0 million of the unremitted foreign earnings. Income taxes have not been provided on approximately \$14.9 million of unremitted foreign earnings because they are considered to be indefinitely reinvested. The tax payable on the earnings that are indefinitely reinvested would be immaterial.
\\\\
\#\#\# Data: \newline\{"header": ["", "Year ended March 31,", ""], "rows": [["", "2019", "2018"], ["Beginning balance", "\$6,164", "\$4,931"], ["Additions based on tax positions related to current year", "164", "142"], ["Additions for tax positions of prior years", "231", "1,444"], ["Reductions due to change in foreign exchange rate  ", "(301)", "(353)"], ["Expiration of statutes of limitation", "(165)", ""], ["Reductions due to settlements with tax authorities", "(77)", ""], ["Ending balance", "\$6,016", "\$6,164"]]\}
\\\\
\#\#\# Response: \newline 164, 142
\\\\
\textbf{Example 3 - The response is an equation}\newline
Below is an instruction that describes a task, coupled with input and data providing additional context. Formulate an arithmetic equation to generate the answer.
\\\\
\#\#\# Instruction:\newline
What is the average value per share that Robert Andersen acquired on vesting?
\\\\
\#\#\# Input:\newline
Option Exercises and Stock Vested The table below sets forth information concerning the number of shares acquired on exercise of option awards and vesting of stock awards in 2019 and the value realized upon vesting by such officers. (1) Amounts realized from the vesting of stock awards are calculated by multiplying the number of shares that vested by the fair market value of a share of our common stock on the vesting date.
\\\\
\#\#\# Data:\newline
\{"header": ["", "Option Awards", "", "Stock Awards", ""], "rows": [["Name", "Number of Shares Acquired on Exercise (\#)", "Value Realized on Exercise (\$)", "Number of Shares Acquired on Vesting (\#)", "Value Realized on Vesting (\$)"], ["Jon Kirchner", "", "", "153,090", "3,428,285"], ["Robert Andersen", "", "", "24,500", "578,806"], ["Paul Davis", "", "", "20,500", "482,680"], ["Murali Dharan", "", "", "15,000", "330,120"], ["Geir Skaaden", "", "", "21,100", "500,804"]]\}
\\\\
\#\#\# Equation:\newline
578,806/24,500

\subsection{PhraseBank}
\textbf{Example 1} \newline
Below is an instruction that describes a task, paired with an input that provides further context. Write a response that appropriately completes the request. 
\\\\
\#\#\# Instruction: \newline Determine the sentiment of the following.
\\\\
\#\#\# Input: \newline The plant will be fired with a combination of spruce bark, chipped logging residues or milled peat. 
\\\\
\#\#\# Response: \newline neutral 
\\\\
\textbf{Example 2} \newline
Below is an instruction that describes a task, paired with an input that provides further context. Write a response that appropriately completes the request. 
\\\\
\#\#\# Instruction: \newline Determine the sentiment of the following. 
\\\\
\#\#\# Input: \newline Operating profit improved by 27\% to EUR 579.8mn from EUR 457.2mn in 2006. 
\\\\
\#\#\# Response: \newline positive


\subsection{Wiki-SQL}
\textbf{Example 1}\newline
Below is an instruction that describes a task, coupled with contextual data. Compose an SQL script capable of being run on the data to generate the solution.
\\\\
\#\#\# Instruction:\newline How many people watched at Glenferrie Oval? 
\\\\
\#\#\# Data:\newline \{"header": ["Home team", "Home team score", "Away team", "Away team score", "Venue", "Crowd", "Date"], "rows": [["North Melbourne", "12.10 (82)", "South Melbourne", "11.14 (80)", "Arden Street Oval", "6,000", "4 August 1928"], ["Fitzroy", "13.12 (90)", "Footscray", "12.17 (89)", "Brunswick Street Oval", "12,000", "4 August 1928"], ["Richmond", "11.13 (79)", "Melbourne", "7.8 (50)", "Punt Road Oval", "26,000", "4 August 1928"], ["Geelong", "4.14 (38)", "Essendon", "12.10 (82)", "Corio Oval", "10,000", "4 August 1928"], ["Hawthorn", "9.9 (63)", "Collingwood", "17.18 (120)", "Glenferrie Oval", "5,000", "4 August 1928"], ["St Kilda", "13.15 (93)", "Carlton", "10.9 (69)", "Junction Oval", "31,000", "4 August 1928"]], "types": ["text", "text", "text", "text", "text", "real", "text"], "caption": "Round 15"\}
\\\\
\#\#\# SQL:\newline SELECT SUM([Crowd]) FROM data\_table WHERE LOWER([Venue]) = LOWER('glenferrie oval')'

\subsection{OTT-QA}
\textbf{Example 1}\newline
Here is a instruction detailing a task, accompanied by data providing additional context. Provide a suitable reply that effectively fulfills the inquiry.
\\\\
\#\#\# Instruction:\newline
How many kilometers is the airport from the Australian city known for housing the Towsers Huts?\\\\
\#\#\# Data:\newline
\{"header": ["Community", "Airport name", "Type", "ICAO", "IATA"], "rows": [["Albury", "Albury Airport", "Public", "YMAY", "ABX"], ["Armidale", "Armidale Airport", "Public", "YARM", "ARM"], ["Ballina", "Ballina Byron Gateway Airport", "Public", "YBNA", "BNK"], ["Balranald", "Balranald Airport", "Public", "YBRN", "BZD"], ["Bankstown , Sydney", "Bankstown Airport", "Airschool", "YSBK", "BWU"], ["Bathurst", "Bathurst Airport", "Public", "YBTH", "BHS"], ["Bourke", "Bourke Airport", "Public", "YBKE", "BRK"], ["Brewarrina", "Brewarrina Airport", "Public", "YBRW", "BWQ"], ["Broken Hill", "Broken Hill Airport", "Public", "YBHI", "BHQ"], ["Camden", "Camden Airport", "Public", "YSCN", "CDU"], ["Cessnock", "Cessnock Airport", "Public", "YCNK", "CES"], ["Cobar", "Cobar Airport", "Public", "YCBA", "CAZ"], ["Coffs Harbour", "Coffs Harbour Airport", "Public", "YCFS", "CFS"], ["Collarenebri", "Collarenebri Airport", "Public", "YCBR", "CRB"], ["Condobolin", "Condobolin Airport", "Public", "YCDO", "CBX"], ["Coolah", "Coolah Airport", "Public", "YCAH", ""], ["Cooma", "Cooma - Polo Flat Airport", "Public", "YPFT", ""], ["Cooma", "Cooma - Snowy Mountains Airport", "Public", "YCOM", "OOM"], ["Coonabarabran", "Coonabarabran Airport", "Public", "YCBB", "COJ"], ["Coonamble", "Coonamble Airport", "Public", "YCNM", "CNB"]], "caption": "List of airports in New South Wales"\}
\\\\
\#\#\# Response:\newline
3

\subsection{Template choice}
\textbf{Example 1 - Arithmetic Template}\newline
Here is a instruction, input and data detailing a task. Which template is best suited to fulfil this inquiry.
\\\\
\#\#\# Instruction:\newline
What was the \% change in gains recognized in other comprehensive income (loss), net of tax of \$1, \$11, and \$4 from 2018 to 2019?
\\\\
\#\#\# Input:\newline
Cash Flow Hedge Gains (Losses) We recognized the following gains (losses) on foreign exchange contracts designated as cash flow hedges: We do not have any net derivative gains included in AOCI as of June 30, 2019 that will be reclassified into earnings within the following 12 months. No significant amounts of gains (losses) were reclassified from AOCI into earnings as a result of forecasted transactions that failed to occur during fiscal year 2019.
\\\\
\#\#\# Data:\newline
\{"header": ["(In millions)", "", "", ""], "rows": [["Year Ended June 30,", "2019", "2018", "2017"], ["Effective Portion", "", "", ""], ["Gains recognized in other comprehensive income (loss), net of tax of \$1, \$11, and \$4", "\$  159", "\$  219", "\$  328"], ["Gains reclassified from accumulated other comprehensive income (loss) into revenue", "341", "185", "555"], ["Amount Excluded from Effectiveness Assessment and Ineffective Portion", "", "", ""], ["Losses recognized in other income (expense), net", "(64)", "(255)", "(389)"]]\}
\\\\
\#\#\# Template:\newline
arithmetic
\\\\
\textbf{Example 2 - Script Template}\newline
Here is a instruction and data detailing a task. Which template is best suited to fulfil this inquiry.
\\\\
\#\#\# Instruction:\newline
In what division was there a population density in km2 of 4,491.8 in 2011?
\\\\
\#\#\# Data:\newline
\{"header": ["Administrative division", "Area (km) 2011**", "Population 2001 Census (Adjusted)", "Population 2011 Census (Adjusted)", "Population density (/km 2011)"], "rows": [["Dhaka District", "1,463.6", 9036647, 12517361, "8,552.4"], ["=> Savar Upazila", "282.11", 629695, 1442885, "5,114.6"], ["=> Keraniganj Upazila", "166.82", 649373, 824538, "4,942.68"], ["Narayanganj District", "684.37", 2300514, 3074078, "4,491.8"], ["=> Narayanganj Sadar Upazila", "100.74", 946205, 1381796, "13,716.5"], ["=> Bandar Upazila", "54.39", 267021, 327149, "6,014.8"], ["=> Rupganj Upazila", "176.48", 423135, 558192, "3,162.9"], ["Gazipur District", "1,806.36", 2143200, 3548115, "1,964.2"], ["=> Gazipur Sadar Upazila", "457.67", 925454, 1899575, "4,150.5"], ["=> Kaliakair Upazila", "314.13", 278967, 503976, "1,604.3"], ["Narsingdi District", "1,150.14", 1983499, 2314899, "2,012.7"], ["=> Narsingdi Sadar Upazila", "213.43", 606474, 737362, "3,454.8"], ["=> Palash Upazila", "94.43", 198106, 221979, "2,350.7"]], "types": ["text", "text", "real", "real", "text"]\}
\\\\
\#\#\# Template:\newline
script

\end{document}